\documentclass[11pt,a4paper]{article}
\usepackage[hyperref]{acl2020}
\usepackage{times}
\usepackage{latexsym}
\usepackage{amsmath}
\usepackage{cancel}
\usepackage{enumitem}
\usepackage{graphicx}
\usepackage{url}
\usepackage{pifont}
\usepackage{xcolor}
\usepackage{soul}
\newcommand{\cmark}{\ding{51}}%
\newcommand{\xmark}{\ding{55}}%

\definecolor{lightgreen}{rgb}{0.56, 0.93, 0.56}
\newcommand{\cfbox}[2]{%
    \colorlet{currentcolor}{.}%
    {\color{#1}%
    \fbox{\color{currentcolor}#2}}%
}

\newcommand{\nt}[1]{\textcolor{blue}{#1}}

\usepackage{microtype}

\aclfinalcopy 
\def\aclpaperid{1390} 

\newcommand{\baseModel}{\textsc{GCC} }
\newcommand{\longModel}{\textsc{Generative Conversation Control} }
\newcommand{\gpt}{\textsc{GPT-2} }
\newcommand{\gptorig}{\textsc{GPT} }
\newcommand{\decModel}{\textsc{GCC-Dec} }
\newcommand{\stsModel}{\textsc{GCC-S2S} }
\newcommand{\vaeModel}{\textsc{GCC-VAE} }
\newcommand{\norefModel}{\textsc{GCC-Nrc} }

\title{Large Scale Multi-Actor Generative Dialog Modeling} 

\author{Alex Boyd\thanks{\; First two authors have contributed equally.} \- \thanks{\; Research conducted during an internship at NVIDIA.}\\
  Department of Statistics \\
  University of California, Irvine \\
  \texttt{alexjb@uci.edu} \\ \And
  Raul Puri\footnote[1]{}\\
  NVIDIA\\
  \texttt{raulp@nvidia.com}  \\\And
  Mohammad Shoeybi\\
  NVIDIA\\
  \texttt{mshoeybi@nvidia.com}  \\\AND
  Mostofa Patwary\\
  NVIDIA\\
  \texttt{mpatwary@nvidia.com}  \\\And
  Bryan Catanzaro\\
  NVIDIA\\
  \texttt{bcatanzaro@nvidia.com} }

\date{}

\begin{document}
\maketitle
\begin{abstract} 
Non-goal oriented dialog agents (i.e. chatbots) aim to produce varying and engaging conversations with a user; however, they typically exhibit either inconsistent personality across conversations or the average personality of all users. 
This paper addresses these issues by controlling an agent's persona upon generation via conditioning on prior conversations of a target actor. In doing so, we are able to utilize more abstract patterns within a person's speech and better emulate them in generated responses. This work introduces the \longModel model, an augmented and fine-tuned \gpt language model that conditions on past reference conversations to probabilistically model multi-turn conversations in the actor's persona. We introduce an accompanying data collection procedure to obtain 10.3M conversations from 6 months worth of Reddit comments. 
We demonstrate that scaling model sizes from 117M to 8.3B parameters yields an improvement from 23.14 to 13.14 perplexity on 1.7M held out Reddit conversations. Increasing model scale yielded similar improvements in human evaluations that measure preference of model samples to the held out target distribution in terms of realism (31\% increased to 37\% preference), style matching (37\% to 42\%), grammar and content quality (29\% to 42\%), and conversation coherency (32\% to 40\%). We find that conditionally modeling past conversations improves perplexity by 0.47 in automatic evaluations. Through human trials we identify positive trends between conditional modeling and style matching and outline steps to further improve persona control. 

\end{abstract}

\begin{table*}[]
    \centering\scalebox{0.835}{
    \begin{tabular}{cp{13.5cm}}
    \hline
    Speaker & Conversation Turn \\
    \hline \hline
    A    & They are worried about themes becoming an exploit. It happened multiple times with the \colorbox{lightgreen}{3DS}\\
    B    & How was themes an exploit on the 3ds \\
    A    & You would inject the theme and holding L or R at boot would start the homebrew launcher \\
    \hline
    \emph{B}    & \emph{Thanks \colorbox{yellow}{i} was not aware of that and thought \colorbox{yellow}{i} would learn a new thing} \\
    \emph{A}    & \emph{I mean, \colorbox{lightgreen}{3DS} has been out for some time} \\
    \emph{B}    & \emph{yeah but \cfbox{blue}{\colorbox{yellow}{i} only started playing it in december 2018}} \\
    \hline
    \multicolumn{2}{c}{(i)} \\
    \hline
    \# & (Ref. Parent Comment) $\rightarrow$ Ref. Reply Comment for Speaker B \\
    \hline \hline
    1 & (\emph{n/a}) $\rightarrow$ \colorbox{yellow}{i} once had 100 pings from the same channel and all was because the owner did not know how to make a long comment \\
    2 & (You're worse than us.) $\rightarrow$ And im one of them but \colorbox{yellow}{i} do got skins im just just as bad \\
    3 & (\emph{n/a}) $\rightarrow$ Oh well \cfbox{blue}{im a newbie} \\
    4 & (\emph{n/a}) $\rightarrow$ oh wow \colorbox{yellow}{i} did not see the read it backwards thing so \colorbox{yellow}{i} did not understand only when \colorbox{yellow}{i} scrolled down \\
    \hline
    \multicolumn{2}{c}{(ii)}
    \end{tabular}}
    \caption{(i) is a conversation sampled from our \decModel(8.3B) model with \nt{(ii)} corresponding to the reference material from speaker B (`\emph{n/a}' indicates no parent comment for the given reference comment). Reference material for speaker A was not included for brevity. The first three turns in (i) are from a real conversation had within the validation set. The last three turns (\emph{italicized}) were generated by sampling turns sequentially, alternating the target speaker between B and A. \colorbox{yellow}{Yellow highlights} indicate where the model appropriately transferred a part of style between references and generations for speaker B (e.g. `i' v. `I'), \cfbox{blue}{boxed words} indicate a transfer of content for speaker B, and \colorbox{lightgreen}{green highlights} indicate consistent style for turns from speaker A (e.g. `3DS' v. `3ds').}
    \label{tab:convo_example_1}
    \vspace{-4mm}
\end{table*}

\section{Introduction} 
Modeling dialog agents, otherwise known as chatbots, has been a longstanding goal within artificial intelligence research. Historically, approaches to this task can be divided into one of the two categories: {\it retrieval} and {\it generative}. The former is posed as a search problem where an appropriate response for a conversation is selected from a large set of candidate replies, whereas the latter autoregressively samples a reply, thereby potentially creating a response that the model may not have seen before. The flexibility and creativity afforded by not prespecifying every possible response is a major draw for generative based approaches. 

In recent years, advances in neural methods have shown promise in effectively modeling this task. Early progress first demonstrated potential with recurrent network based models capable of holding simple conversations \citep{sordoni2015neural}. Further architecture optimizations and tweaks improved user experiences; however, they largely experienced issues with the agent exhibiting an inconsistent personality and producing uninteresting comments \citep{li2015diversity}. Some works have attempted to alleviate this through conditioning on various factors of the conversation through methods such as sentiment or speaker embeddings \citep{li2016persona}, but the added data annotation makes these methods not scale well to the gargantuan amounts of data needed to train larger models. 

A persona-based conversation task was introduced by \citet{zhang2018personalizing} where a set of Reddit comments and their replies were accompanied by brief descriptions or factoids about the speakers, such as their hobbies and interests. Recent works \citet{wolf2019transfertransfo} have shown that leveraging this format with pre-trained transformer-based language models yield state-of-the-art (SOTA) performance in generative conversation modeling. However, in our interactions with these models they produced conversations that adhered to the reference facts, but were devoid of unique personality and instead exhibited a mean average style.

Personality, as seen through text, manifests itself not just through content, but also through a person's tone, grammar, and vernacular. As such, a criticism of prior persona-based solutions is that the ``personas'' only reflect surface-level characteristics of a person's manner of speaking and can result in banal generations. What \emph{does} showcase a personality are actual conversation examples from a person. By conditioning on previous, unrelated, conversation turns for a speaker, we generate new replies that utilize more abstract personality traits inherent in the reference examples. We define this as a \emph{conditional conversation} task.

Emulating this abstract notion of style requires large amount of data and sufficiently powerful model. We propose a data collection procedure that heuristically scrapes user data and comment threads from Reddit\footnote{\url{https://reddit.com/}} to produce conversations that vary widely in content, speakers, and reference histories to condition on. This work also introduces the \longModel  (\baseModel\kern-0.2em) model, an augmented and fine-tuned \gpt language model. We take advantage of large transformers' ability to model long contexts and dependencies, and successfully model multi-turn and multi-actor conversations that are significantly longer (up to 15 turns) than most prior work.

We find that scaling model sizes from 117M to 8.3B parameters yields an improvement from 23.14 to 13.14 perplexity on 1.7M held out Reddit conversations. Similar improvements from model scaling are found in human evaluations that measure sample preference to the held out target distribution in terms of realism (31\% increased to 37\% preference), style matching (37\% to 42\%), grammar and content quality (29\% to 42\%), and conversation coherency (32\% to 40\%).

To summarize, our contributions in this work are three-fold:

\begin{enumerate}[label=\roman*]
    \itemsep-0.3em 
    \item We introduce a new conversational task and demonstrate added value over traditional conversation modeling through both better control and response generation.
    \item We document the creation of a large, multi-turn, multi-actor conversational dataset and the techniques used to clean it and extract conversations and reference material for style. 
    \item We demonstrate that by increasing model size from 117M to 8.3B parameters, human evaluations measuring preference of model generated samples over held out target distribution increase with respect to realism, style matching, grammar, and conversation coherency. Automatic evaluations also showcase similar trends with the largest model leading to significantly lower perplexities.
\end{enumerate}

\section{Conversation Modeling}

Let $\mathbf{c}$ represent a multi-turn conversation of variable-length, and let $\mathbf{x}_j$ represent a single turn that contains a variable-amount of tokens.
Mathematically, this is represented as $\mathbf{c} = (\mathbf{x}_1, \dots, \mathbf{x}_{|\mathbf{c}|})$, with $\mathbf{x}_j = (x_{j,1}, \dots, x_{j,|\mathbf{x}_j|})$.
Every token, in every turn, belongs to the same fixed vocabulary (i.e. $x_{j,t} \in \mathcal{V}$). 
Assume that $p_*(\cdot)$ represents the true distribution of content.

\subsection{Language Modeling and Dialog}

Standard language modeling involves modeling sequences of tokens. 
After factorizing, the problem is most commonly construed as a next-token prediction problem where $p_*(\mathbf{x})$ is approximated via:
\begin{equation}
    p_\theta(\mathbf{x}) = \prod_{t=1}^{|\mathbf{x}|} p_\theta(x_t | x_{<t})
\end{equation}

where $\theta$ is optimized over a set of documents, $\mathcal{D} = \{\mathbf{x}^{(1)}, \dots, \mathbf{x}^{|\mathcal{D}|}\}$, using maximum likelihood estimation:
\begin{equation}
    \mathcal{L}(\theta, \mathcal{D}) = \sum_{i=1}^{|\mathcal{D}|} \log p_\theta(\mathbf{x}^{(i)})
\end{equation}

Likewise, to model dialog in the same vein requires just a small alteration. Instead of modeling just a single sequence of tokens, $\mathbf{x}$, the new objective is to model several sequences of tokens that comprise a conversation, $\mathbf{c}$. As such, $p_*(\mathbf{c})$ is approximated via:
\begin{equation}\begin{split}
    p_\theta(\mathbf{c}) & = \prod_{j=1}^{|\mathbf{c}|} p_\theta(\mathbf{x}_j | \mathbf{x}_{<j}) \\
    & = \prod_{j=1}^{|\mathbf{c}|} \prod_{t=1}^{|\mathbf{x}_j|} p_\theta(x_{j,t} | x_{j,<t}, \mathbf{x}_{<j}) \label{eq:1}
\end{split}\end{equation}

where $\theta$ is optimized over a set of conversations, $\mathcal{D} = \{\mathbf{c}^{(1)}, \dots, \mathbf{c}^{|\mathcal{D}|}\}$, using maximum likelihood estimation:
\begin{equation}
    \mathcal{L}(\theta, \mathcal{D}) = \sum_{i=1}^{|\mathcal{D}|} \log p_\theta(\mathbf{c}^{(i)})
\end{equation}

\subsection{Conditioning on Prior Conversations}

To have more control over generation and better insight into the distribution of turns within a conversation, it is better to conditionally model $\mathbf{c}$ instead of modeling it unconditionally as in Equation \ref{eq:1}. 

For every turn in a particular conversation, $\mathbf{x}_j \in \mathbf{c}$, let $\mathbf{r}_j$ be a corresponding set of reference history tuples. These tuples contain (i) a prior turn of conversation (ii) a turn of conversation spoken by the same agent as $\mathbf{x}_j$ in response to the first member of the tuple. In the event that (ii) corresponds to the beginning of a conversation (i) is left blank. We stipulate that the turns of $\mathbf{c}$ and the turns of $\mathbf{r}_j$ are disjoint. This is defined mathematically as: 
\begin{equation}\begin{split}
    \mathbf{r}_j = \{(\mathbf{x}_{k-1}, \mathbf{x}_{k}) \;|\;& \text{author}(\mathbf{x}_{k}) = \text{author}(\mathbf{x}_{j}) \\
    & \land \mathbf{x}_k \notin \mathbf{c}\}
\end{split}\end{equation}

The intention of including previous replies by the same person is to get a better idea of the personality, tone, vernacular, and content of potential responses when predicting next tokens for the given turn. Likewise, the turns that the agent was replying to in $\mathbf{r}_j$ are also included to get a better idea as to the transition dynamics of how they interact with other agents. 

We update our prior equations to reflect this change in modeling objective:
\begin{equation}\begin{split}
    p_\theta(\mathbf{c} | \mathbf{r}) & = \prod_{j=1}^{|\mathbf{c}|} p_\theta(\mathbf{x}_j | \mathbf{x}_{<j}, \mathbf{r}_j) \\
    & = \prod_{j=1}^{|\mathbf{c}|} \prod_{t=1}^{|\mathbf{x}_j|} p_\theta(x_{j,t} | x_{j,<t}, \mathbf{x}_{<j}, \mathbf{r}_j) \\
    \mathcal{L}(\theta, \mathcal{D}) & = \sum_{i=1}^{|\mathcal{D}|} \log p_\theta(\mathbf{c}^{(i)} | \mathbf{r}^{(i)})
\end{split}\end{equation}

\section{Data}

In order to sufficiently train a model to be able to autoregressively generate turns in a conversation conditioned on prior conversations, we require an ample amount of diverse examples accompanied with plenty of reference material. A suitable source of data for this purpose can be found from comments made on Reddit posts. Thanks to a publicly available archive on \url{pushshift.io}, comments are processed from Reddit ranging from October of 2018 to March of 2019 for training, and April of 2019 for validation. The techniques described in this section can naturally be extended to the full range of Reddit data spanning as far back as 2005; however, we choose to focus on just the 6 months in question for the sake of tractability.
 
\subsection{Extracting Conversations}

Comments for a singular post on Reddit naturally exist as a tree structure; however, conversations necessitate a sequence of turns. As such, we obtain conversations by extracting ``valid"  paths from the comment graph structure. Paths are extracted sequentially from the longest candidates to shortest, and a candidate path is considered valid if and only if it satisfy the following conditions:

\begin{enumerate}
    \itemsep-0.3em 
    \item The path has a minimum of 5 turns
    \item The path has a maximum of 15 turns
    \item At least one turn has minimum karma score\footnote{``Karma" can be thought of as the net amount of likes and dislikes a comment has, as voted upon by the users of Reddit.} of 4 within the path
    \item All turns in the path have at least 3 words
    \item The path shares a maximum of 2 turns with previously extracted paths
    \item No turns in the path originate from a ``not safe for work" subreddit
\end{enumerate}

These rules were decided upon to ensure that the model is able to learn multi-turn conversations (rules 1 and 2) with appropriate and meaningful comments being made (3, 4, and 6) while ensuring a diverse set of examples (5) are available. 

Due to memory constraints, comments are only processed on a month to month basis so any conversations that span across months are lost; however, this loss is negligible due to the vast amount of data at hand. Furthermore, this technique possibly results in more relevant references than those collected from prior months as the reference data is temporally local to the conversations in question and reflects users' current personas and interests.

After all conversations have been extracted, a reference set of turns (and comments that they were replying to) are collected for every user. We save, at most if available, the top 8 scoring comments for every user. Most users have much more than 8 comments, so an average of 7.1 reference tuples per user are collected, with about half of the tuples containing a parent comment that the user was replying to.

\begin{table}[]
    \centering\scalebox{0.835}{
    \begin{tabular}{ccccc}
    \hline
    Dataset & Convos & Turns & Users & Refs \\
    \hline \hline
    Training &  10.3M & 72.6M & 10.5M & 73.4M \\
    Validation & 1.8M & 12.5M & 1.9M & 13.2M \\
    \hline
    \end{tabular}}
    \caption{Statistics of the training and validation datasets detailing the amount of different conversations, turns, users, and reference tuples present in each. The training and validation sets were processed from Reddit comment threads spanning from October, 2018 to March, 2019 and April, 2019 respectively.}
    \label{tab:dataset}
\vspace{-4mm}    
\end{table}

\section{Model}
\begin{figure*}[th!]
    \begin{center}
    \scalebox{0.87}{
      \includegraphics[width=\linewidth,height=11em]{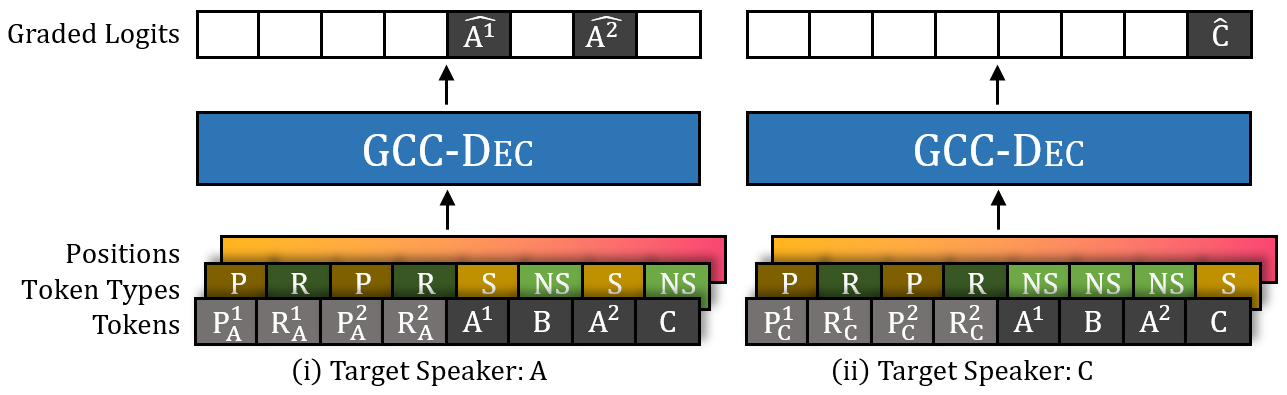}}
      \caption{Illustration of input representation for a conversation from three different speakers $(A,B,C)$ composed of a sequence of four turns (denoted $A^1, B, A^2, C$ in the ``Tokens" row) when separately modeling two different target speakers ($A$ and $C$ for (i) and (ii) respectively). The model receives different reference histories (parent comments $P^j_{A/C}$ and associated reply $R^j_{A/C}$ from target speaker $A/C$ for $j\in \{1,2\}$) and different placements of token types based on which Target Speaker is modeled (in ``Token Types" row, $P$ denotes parent comment, $R$ is reply comment, $S$ is a turn from target speaker, and $NS$ is a turn not from the target speaker. Losses are computed only for the comments corresponding to the active Target Speaker ($\hat{A^j}$ v. $\hat{C}$). Note that this representation explicitly allows for multi-actor conversation modeling.}
      \label{fig:input}
  \end{center}
\vspace{-4mm}
\end{figure*}

All models proposed stem from the \gpt model architecture as their base design \citep{radford2019language}. The class of models will be defined as \longModel models, \baseModel We experiment with the number of layers, $l$, the hidden size, $h$, and the number of attention heads, $A$ in the \gpt model architecture.
Modeling conversations with \baseModel requires three steps: (i) identify a speaker to emulate and obtain their reference history consisting of comments they made on other Reddit posts, (ii) input the reference history and conversation turns into the model, and (iii) retrieve estimated next-token probabilities only associated with turns in the conversation spoken by the target speaker.

\subsection{Data Representation}

Due to supporting multi-actor conversations present in our dataset, special care is needed for presenting this information to the model. In general, this is accomplished by designating a speaker of interest to model in a conversation. 

As visualized in Figure \ref{fig:input}, the designated speaker's reference history tokens are gathered and concatenated together, with a parent comment followed by its associated reply (made by the speaker of interest) followed by another parent comment and so forth. Positional embeddings will signal to the model the order of comments being made; however, additional signal is needed to tell the model which comments are made by the speaker of interest and which are not. This is achieved by token type embeddings that get added to the positional and vocabulary embeddings. All tokens in the reference history that belong to the speaker get the same token type embedding, and all others get a different one. This representation choice allows us to naturally handle multi-actor conversation by only making a distinction between the speaking user and non speaking users. Reference history sequences larger than 512 are truncated from the end to keep the length within 512 tokens.

The conversation turns are similarly represented by concatenating them together in order of occurrence with a special token at the beginning signifying the start of the conversation. For practicality, all turns after the final turn associated with the target speaker are discarded for a given iteration. Each token in the conversation sequence receives a specific token type embedding if it is associated with the speaker of interest, and receives a different type if not. Note, the conversation and reference history have disjoint sets of token type embeddings to differentiate the different types of content. The max length a conversation sequence can be is 512 tokens with extra tokens truncated from the beginning of the conversation to encourage a variety of conversation lengths. In models that have access to the reference history this leads to a total sequence length of 1024 tokens and 512 tokens otherwise.

\subsection{Architectures}

There is flexibility in how to model conversations with reference histories due to the turns in a conversation and reference comments being indirectly related, both content and style-wise. As such, the design choices we consider either encode the references separate from the conversation, or together.

\paragraph{Decoder-Only: \decModel} The simplest of the three considered models consists of only a transformer for decoding, which is the original configuration for \gpt. The input consists of the reference history tokens concatenated with the conversation turn tokens and the corresponding token types. A left-to-right (LR) mask is used across the entire sequence. See Figure~\ref{fig:input} for an illustration. Despite it's simplicity we find that this model performs the best.
 
\paragraph{Seq2Seq Baseline: \stsModel} For this model the reference material with corresponding token types is encoded in a separate transformer using a bidirectional mask. The conversation turns are then decoded with a LR mask using both self-attention and attention against the final hidden states of the encoded reference. This is representative of the typical formulation for attention-based Seq2Seq models \citep{vaswani2017attention}.

\paragraph{Variational Autoencoder Baseline: \vaeModel} This configuration also encodes the reference history and corresponding token types in a separate transformer using a bidirectional mask. The final hidden state of a special classification token is then linearly transformed into the sufficient statistics of a normal latent state which is then sampled. This latent state is then prepended to the embedded inputs of the conversation turns. The final sequence is then decoded using a LR mask in a separate transformer. We explored this method as latent variables are commonly used to control aspects of style across various areas of study.

\paragraph{No Reference Context Baseline: \norefModel} This version is similar to \decModel except that there are no reference material included when decoding information. This model can be seen as a re-implementation of \citet{olabiyi2019multi} with the minor differences being that we introduced token types for multi-actor modeling and we did not utilize their random padding strategy. We found this unnecessary as we did not experience overfitting due to the large amount of training data available. As such, \norefModel will largely serve as our previous SOTA baseline to compare against when demonstrating the advantage of conditioning on prior conversations.

\section{Experiments}

It is known that for the language modeling validation perplexity measures using teacher forcing is not the best evaluation of generative capabilities, even if there is correlation between the two. However, it is a commonly used metric for language modeling, and can be parallelized and computed inexpensively without the need for autoregressive sampling of output text. With that in mind, two sets of evaluations were done, the first of which being an architecture search using automatic evaluation with validation perplexity and the second being a qualitative study using Amazon's Mechanical Turk\footnote{\url{https://www.mturk.com/}} to assess generated samples.

\subsection{Automatic Evaluation} 

All evaluations in this section are done on the validation set (Reddit comments from April, 2019) using perplexity, which is calculated as follows:

\begin{equation}
    \text{PPL}(\theta, \mathcal{D}) = \exp\left\{-\frac{1}{|\mathcal{D}|}\sum_{i=1}^{|\mathcal{D}|} \log p_\theta(\mathbf{c}^{(i)} | \mathbf{r}^{(i)}) \right\}
\end{equation}

All models are trained using mixed precision arithmetic, a learning rate that linearly increases from $0.0$ to $1.5e-4$ over the first 1\% of iterations followed by it decaying to $0.0$ over the remaining iterations with a cosine annealing schedule, and the Adam optimization algorithm with default hyperparameters \citep{kingma2014adam}.

\begin{table}[!t]
\centering\scalebox{0.835}{
\begin{tabular}{cccccc}
\hline 
Model      & $h$  & $l$ & $A$ & Params & PPL \\ 
\hline\hline
\stsModel  & 768  & 18  & 16  & 375M   & 22.09 \\
\vaeModel  & 768  & 20  & 16  & 362M   & 22.43\\
\hline\hline
\decModel  & 1024 & 24  & 16  & 355M   & \textbf{19.10}\\
\hline\hline
\stsModel  & 1024 & 24  & 16  & 810M   & 19.89 \\
\vaeModel  & 1024 & 24  & 16  & 711M   & 20.49 \\
\hline
\end{tabular}}
\caption{\label{table:arch} Comparison of model architecture perplexity (PPL) trained from scratch for 200K iterations. The top half of the table are iso-parameter count experiments while the bottom half are iso-architecture experiments. Note that $h$, $l$, and $A$ define the sizes of the encoder and decoder transformers \emph{individually}.}
\vspace{-4mm}
\end{table}

\paragraph{Architecture} We evaluate three main architectures under two scenarios: similar total number of encoder and decoder parameters, and similar total number of decoder parameters. As such, a 355M parameter version of \decModel is compared to two versions each of \stsModel and \vaeModel. 
When present, the encoder and decoder transformers shared the same hidden sizes, number of layers, and number of attention heads.
Additionally, all models were trained from scratch for 200,000 iterations at a global batch size of 256. 

The results are presented in Table \ref{table:arch}. We see that for models with similar parameter counts the \decModel has the advantage, and that under similar decoder sizes having direct access to the reference material (i.e. processing the reference and conversation together in a single decoder) results in superior performance. This indicates that the added complexity from additional encoding is not needed and that concatenating all relevant context is both the simplest, and most effective means of incorporating previous information. Since the parameters are shared and no latent variable bottleneck is used, the model has full access to the information from the references. With this, the self attention operation is able to automatically modify the model's output distribution in complex, non-linear ways without the need for manual architecture design.

\paragraph{Pre-training and References}

We will use \decModel going forward. It is important to see if we can gain additional predictive power using pre-trained models trained on large, diverse, language modeling corpora, or at the very least utilize less computing resources to achieve similar performance. The \decModel trained from scratch in the previous section will be compared against another model of the same size that was pre-trained using Megatron-LM \citep{shoeybi2019megatron}. The pre-trained \decModel will be fine-tuned for 70,000 iterations at a global batch size of 128. We will also compare against \norefModel fine-tuned from the same checkpoint with the same batch size and amount of iterations.

The results can be seen in Table \ref{table:pt}. We observe that with less data, the pre-trained model quickly eclipses the model trained from scratch and achieves better perplexity, highlighting the need for models with robust linguistic features learned from non-Reddit corpora. Additionally, including reference history improves performance as well. This difference of 0.47, while smaller than differences between results from different model sizes, is notable due to the large amount of out of sample data that the models were tested on.

\begin{table}[!t]
\centering\scalebox{0.835}{
\begin{tabular}{ccccc}
\hline 
Model & P.T. & Iter. & Batch & PPL \\ 
\hline\hline
\decModel   & \xmark & 200K & 256 & 19.10 \\
\decModel   & \cmark & 70K  & 128 & \textbf{18.92} \\
\norefModel & \cmark & 70K  & 128 & 19.39 \\
\hline
\end{tabular}}
\caption{\label{table:pt} Comparison of models with and without pre-trained initializations. All models in this study have 355M parameters with $h=1024$, $l=24$, and $A=16$.}
\end{table}

\begin{table}[!t]
\centering\scalebox{0.835}{
\begin{tabular}{cccccc}
\hline 
Model      & $h$  & $l$ & $A$ & Params & PPL \\ 
\hline\hline
\decModel  & 768  & 12  & 12  & 117M   & 23.14 \\
\decModel  & 1024 & 24  & 16  & 355M   & 18.92 \\
\decModel  & 1280 & 36  & 16  & 774M   & 17.18 \\
\decModel  & 1536 & 40  & 16  & 1.2B   & 16.08 \\
\decModel  & 3072 & 72  & 24  & 8.3B   & \textbf{13.24} \\
\hline
\end{tabular}}
\caption{\label{table:size} Comparison of model performance as size varies. All models are trained from a pre-trained checkpoint for 70K iterations with a batch size of 128.}
\vspace{-4mm}
\end{table}

\paragraph{Model Size}

Finally, we performed an ablation study on the size of \decModel used. The different size configurations and results can be seen in Table \ref{table:size}. All models fine-tuned from a pre-trained checkpoint for 70,000 iterations at a global batch size of 128. As shown in \citet{shoeybi2019megatron},  perplexity decreases as the size of the model increases. This increase in performance is significant as it has been shown that for conversational models there is a correlation between held-out perplexity measures and human-likeness of sampled turns, especially for models within the same family \cite{adiwardana2020towards}.

\begin{table*}[!htb]
\centering\scalebox{0.835}{
\begin{tabular}{lccccr}
\hline 
Source A & Realistic & Reference & Quality & Coherency & Source B \\ 
\hline\hline
\norefModel (355M) & 31\%\;-\;35\% & 37\%\;-\;41\% & 29\%\;-\;36\% & 32\%\;-\;39\% & Human \\
\decModel   (355M) & 32\%\;-\;34\% & 38\%\;-\;40\% & 31\%\;-\;33\% & 32\%\;-\;36\% & Human \\
\decModel   (774M) & 31\%\;-\;35\% & 40\%\;-\;39\% & 33\%\;-\;33\% & 34\%\;-\;36\% & Human \\
\decModel   (1.2B) & 32\%\;-\;37\% & 40\%\;-\;40\% & 34\%\;-\;38\% & 29\%\;-\;36\% & Human \\
\decModel   (8.3B) & \textbf{37\%\;-\;40\%} & \textbf{42\%\;-\;38\%} & \textbf{42\%\;-\;42\%} & \textbf{40\%\;-\;42\%} & Human \\
\hline
\hline
\decModel   (355M) & 31\%\;-\;34\% & 41\%\;-\;39\% & 37\%\;-\;36\% & 33\%\;-\;35\% & \norefModel (355M) \\
\decModel   (774M) & 33\%\;-\;33\% & 39\%\;-\;40\% & 34\%\;-\;29\% & 34\%\;-\;36\% & \decModel   (355M) \\
\decModel   (1.2B) & 31\%\;-\;31\% & 40\%\;-\;38\% & 33\%\;-\;32\% & 38\%\;-\;38\% & \decModel   (774M) \\
\decModel   (8.3B) & 41\%\;-\;37\% & 39\%\;-\;43\% & 38\%\;-\;38\% & 42\%\;-\;39\% & \decModel   (1.2B) \\
\hline
\end{tabular}}
\caption{\label{table:human} Experiment results for pairwise comparisons grading if conversation samples seemed human-like (Realistic), were inline with the reference history (Reference), were interesting and had good grammar (Quality), and if they fit the conversation as a whole (Coherency). Percentages reported in the format ``A\% - B\%" indicate how often users reported that samples from source A were better than samples from source B for a given category, and vice versa. Percentage pairs do not sum to 100\% due to users being able to report both samples as being of equal standing, thus resulting in a third (omitted) percentage representing neutral opinions.}
\vspace{-4mm}
\end{table*}

\subsection{Human Evaluation}

The goal of the human evaluations is to verify the results of the quantitative ablations studies concerning both model size and presence of reference history. This is done by presenting participants on Mechanical Turk with 375 different ground truth conversations of variable lengths (2, 4, and 8 turns) in even proportions. We utilize 3 raters per example in our setting. To filter out spurious raters we explicitly detail in the instructions that payment is contingent on spending at least a certain amount of time on the samples and completing a survey about their Reddit use. If a rater fails to satisfy both these conditions we discard their label. Adopting this simple heuristic for rater quality led to the disqualification of 33.2\% of our labels. As is common in other work a single conversation is presented with two different realizations for the last turn \citep{serban2017hierarchical}. These last turns can be either machine-generated or ground truth depending on the experiment; however, every model generates exactly one reply for each of the ground truth to be used across all experiments. Samples where three new turns are generated can be seen in Table \ref{tab:convo_example_1} or in Tables \ref{tab:convo_example_2} - \ref{tab:convo_example_8} in the Appendix.

When presented with these different realizations, the participant is asked to rate the pair on several qualities such as which is likely to be human generated, which follows the references well, which has good quality, and which exhibits good turn-to-turn coherency. For each of these the rater is asked to decide which in the pair showcases these qualities better. Note that the rater has the option of selecting both of them exhibit the quality of interest, or neither of them do. These were conducted in pairs to provide a frame of reference for the rater. We present the findings as paired results to account for grounding effects. Exact phrasings of these questions, several sample conversations, and details on our Turk setup can be found in Appendix \ref{turk_setup}. We found inter-rater agreement in our studies about 75-80\% of the time between 2 of the 3 users who judged samples, and about 10\% of the time all 3 agreed unanimously. This is in light of 4 possible choices and 3 raters. It should be noted that our goal is not to make the distribution between model and human statistically different, but rather to make them as close as possible. We have taken several steps to assure the quality of our human evaluations as mentioned in the previous paragraph. Beyond that, any experiment with sufficient statistical power would need a prohibitively expensive number of samples per comparison.

The results of this study can be seen in Table \ref{table:human}. We find that in pairwise comparisons bigger models nearly always outperform their smaller counterpart across all tests we ran. For our pairwise tests we only considered pairings between a model and the next largest model size due to the prohibitive cost of computing all pairwise comparisons. For tests against our ground truth we found the results to be rather noisy. Generally, we observed that the models were close to 30-40\% in all categories meaning that they were sufficiently similar to the ground truth distribution of data (the neutral options were chosen more frequently). However, we found that our 8.3B parameter model was significantly more polarizing than the rest. The model was capable of generating unique and engaging conversations that, when compared to the ground truth, led to it being explicitly preferred more than other models in all tests. It proved to adhere to the persona more than even the ground truth conversations. In addition to effectively utilizing its references to modulate style as we'd hoped, we also found that its realism, linguistic quality, and coherency was superb. Furthermore, we also tested pairwise comparisons between samples from successive model sizes. On average, the larger model tended to achieve similar or superior performance in all of the categories. All in all, these findings reinforce the results from the quantitative experiments in that larger models better match the target distribution. 

\vspace{-2mm}

\begin{figure}
    \begin{center}
    \scalebox{0.87}{
      \includegraphics[width=\linewidth]{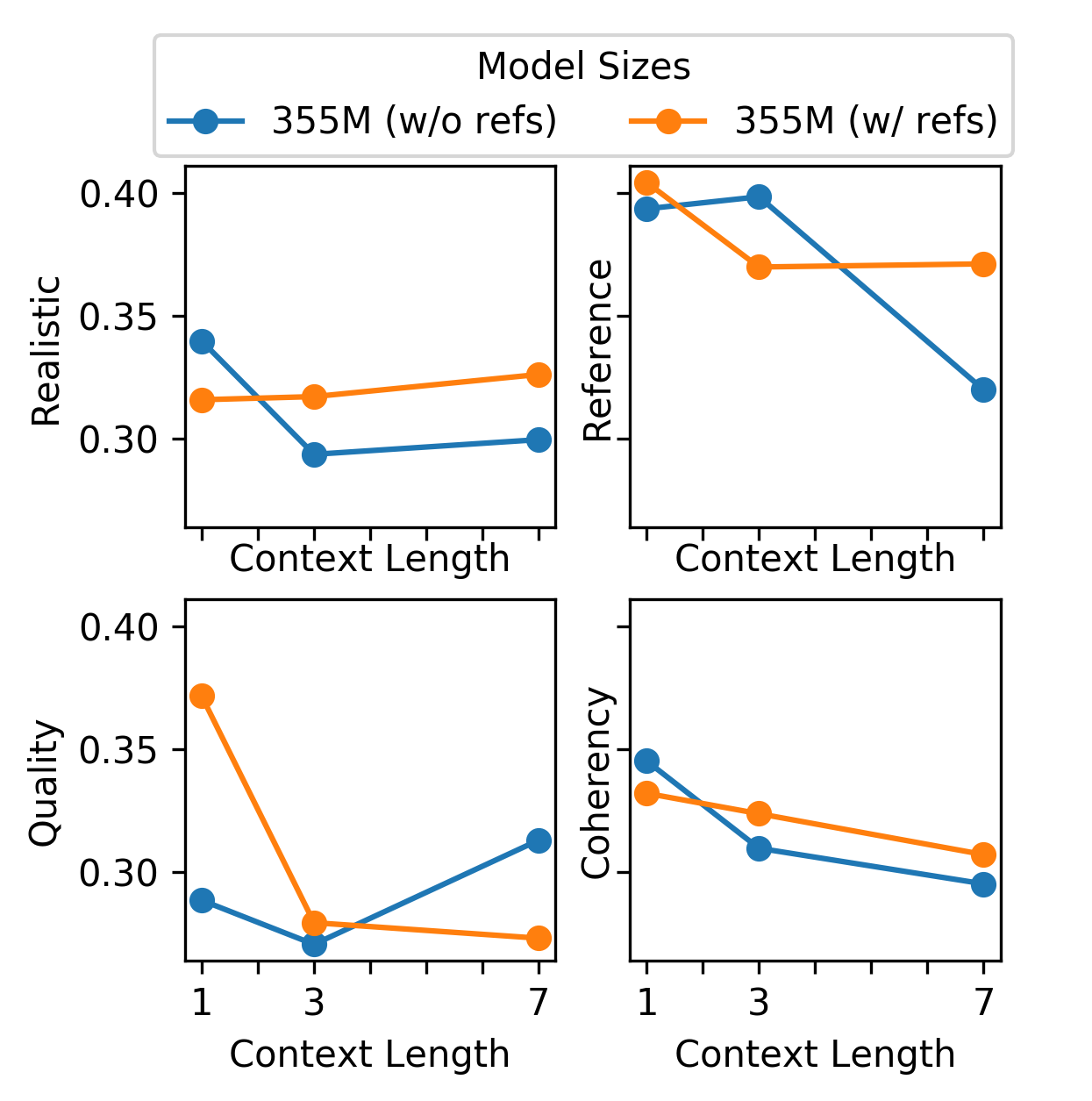}}
      \vspace{-4mm}
      \caption{Test scores compared against the number of dialog turns given as context prior to generating samples for \decModel(355M) and \norefModel(355M).}
      \label{fig:lengths}
  \end{center}
  \vspace{-6mm}
\end{figure}

\paragraph{Reference use}
From our qualitative study we can clearly see the benefit of using reference history as was alluded to in prior sections. In all four experiments the presence of references leads to better ground truth performance compared to \norefModel. In Figure~\ref{fig:lengths} we delve deeper into the results of the ground truth experiments and display labeler preference as a function of conversation length. As can be seen, when the conversation has built up a lot of context, \norefModel (355M) moves away from the user style, instead focusing presumably on the style within the conversation. Alternatively, \decModel   (355M) adheres more closely to the references instead of the prior conversation context, thus resulting in higher style match for longer conversations. However, this over-adherance to the conversation style does seem to impact conversation quality for longer conversations. It is possible that our inclusion of random reference conversations leads to this quality degradation. To investigate this future work could consider incorporating information retrieval components to select contextually relevant reference conversations for more accurate personality transfer that does not degrade conversation quality.

\vspace{-3mm}

\section{Related Work}

\paragraph{Transformer Language Models} \citet{radford2018improving} released the first widely used transformer based generative language model, \gptorig. Follow up work, \gpt, showed that language modeling quality improved as model size grew, up to 1.5B parameters \citep{radford2019language}, and that large transformer language models were able to successfully incorporate long term dependencies to model and generate diverse content. Further work with generative transformer language models would go on to push model scale by testing up to 8.3B parameters and 11B parameters in two separate studies \citep{shoeybi2019megatron,raffel2019exploring}. These results have demonstrated performance scaling not only for the original language modeling task, but also on plenty of downstream NLP tasks as well \citep{radford2018improving,radford2019language,dai2019transformer,howard2018universal,liu2019multi,zellers2019defending,yang2019xlnet,devlin2018bert}. We demonstrate that this scaling trend applies to the conditional conversation modeling task as well and validate the efficacy of transformer based language models for dialog modeling.

\paragraph{Dialog Modeling} Generative, non-goal oriented dialog modeling (i.e. chit-chat) has a history of difficulty with modeling long contexts \citep{serban2016generative}, exhibiting a consistent personality \citep{li2016persona}, and producing interesting and engaging responses \citep{li2015diversity}. In general approaches to mitigating these issues have included: tweaking the base recurrent network architecture to introduce persona-based latent variables (that are either learned, amortized, or adversarially generated) \citep{serban2017hierarchical,bak2019variational,chan2019modeling,olabiyi2019adversarial}, learning speaker embeddings to modulate style \citep{li2016persona}, and conditioning on outside information or heuristics to control generation
\citep{young2018augmenting,joshi2017personalization,ghazvininejad2018knowledge}. One particular way that inconsistent personalities have been addressed is by conditioning the model on a set of sentences describing the target personality \citep{zhang2018personalizing,mazare2018training}. As described in the prior section large transformer models have demonstrated success in generating diverse and engaging content. Recent work in conversational modeling has built upon the success of these transformer-based architectures to allow for longer contexts and incorporating multiple turns \citep{wolf2019transfertransfo,olabiyi2019multi}. 

Several datasets have been proposed for multi-turn conversation modeling \citep{moviedata,ubuntudata}; however, these are limited to relatively short median conversation lengths of 3 and 6-turn respectively. Contexts of these lengths are not able to take full advantage of \gpt and other large transformer's modeling capabilities. Addressing this shortcoming and curating a dataset of diverse conversations that cover a wider distribution of conversation lengths from 0 to 15 turn contexts is a central goal of this work. Concurrent work has shown the value of leveraging large amounts of Reddit data to harvest naturally occurring conversations for the purposes of downstream conversational tasks \cite{zhang2019dialogpt}. However, this work does not address the issue of stylistic control or the effects of scaling models to large sizes, which are central themes of our work. Other concurrent work has also shown the benefit of learning from large amounts of social media conversations, but it also did not attempt to influence the model output style nor did it scale up the model to 8.3 billion parameters \cite{adiwardana2020towards}.

\section{Conclusions}

When a large conversational model is trained on a diverse collection of multi-turn conversations, it is able to generate quality conversations that are engaging, coherent, and plausibly human. Furthermore, when conditioned on prior conversations, the model is able to utilize a speaker's personality when choosing how to reply in a conversation to allow for greater control and more diverse responses. In the future, we aim to leverage these pre-trained models to advance SOTA on downstream conversational tasks, such as knowledge-grounded conversations or question answering. Recent advancements in learnable information retrieval systems could select contextually relevant references to further strengthen the quality of generated dialogue.

\bibliography{acl2020}
\bibliographystyle{acl_natbib}

\appendix

\section{Mechanical Turk Setup}
\label{turk_setup}
\begin{figure*}[h!]
	\begin{center}
		\includegraphics[scale=.4]{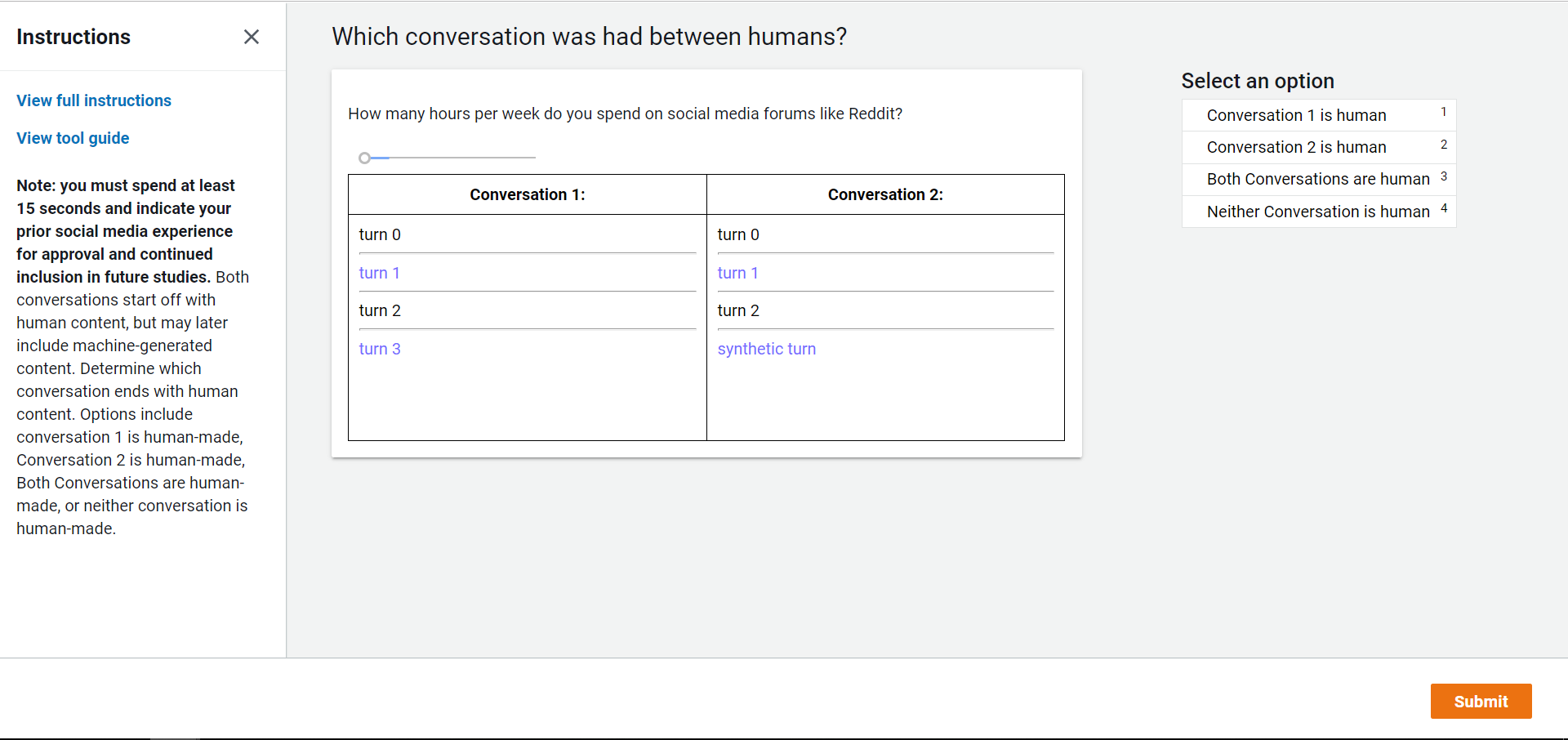}
		\caption{Realism Task Turk Layout.}
		\label{fig:det_turk_layout}
	\end{center}
\end{figure*}

\begin{figure*}[h!]
	\begin{center}
		\includegraphics[scale=.4]{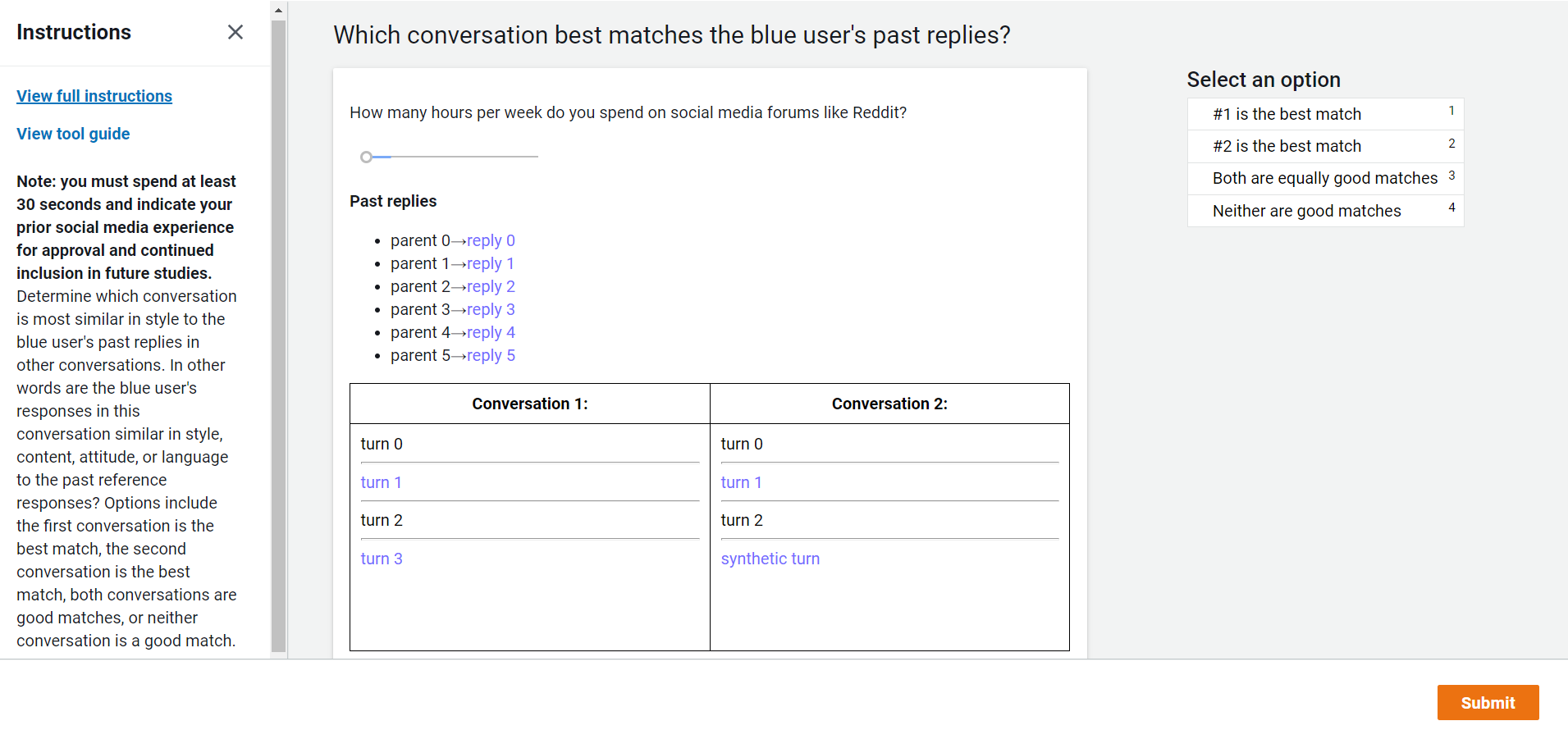}
		\caption{Reference Task Turk Layout.}
		\label{fig:ref_turk_layout}
	\end{center}
\end{figure*}

\begin{figure*}[h!]
	\begin{center}
		\includegraphics[scale=.4]{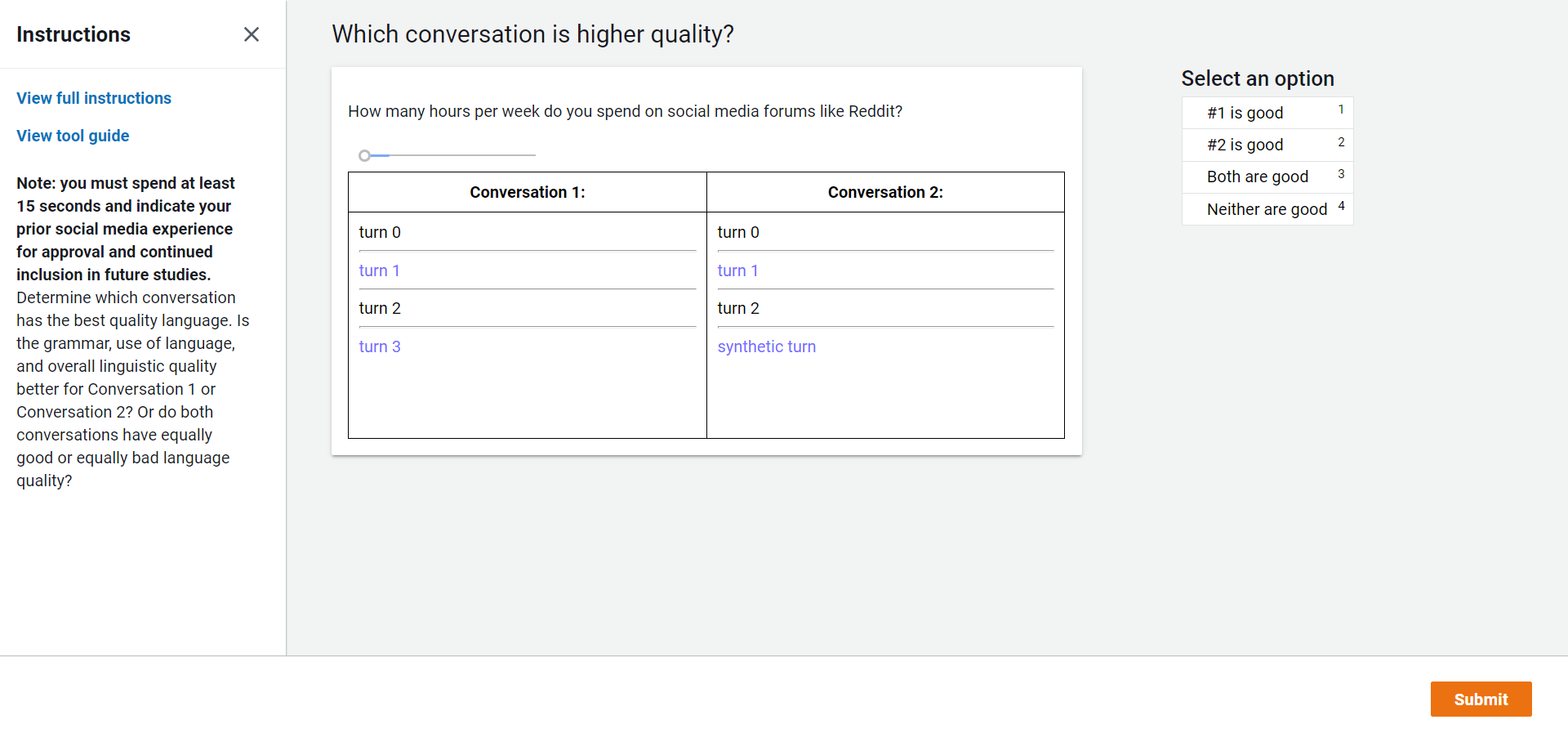}
		\caption{Quality Task Turk Layout.}
		\label{fig:qua_turk_layout}
	\end{center}
\end{figure*}

\begin{figure*}[h!]
	\begin{center}
		\includegraphics[scale=.4]{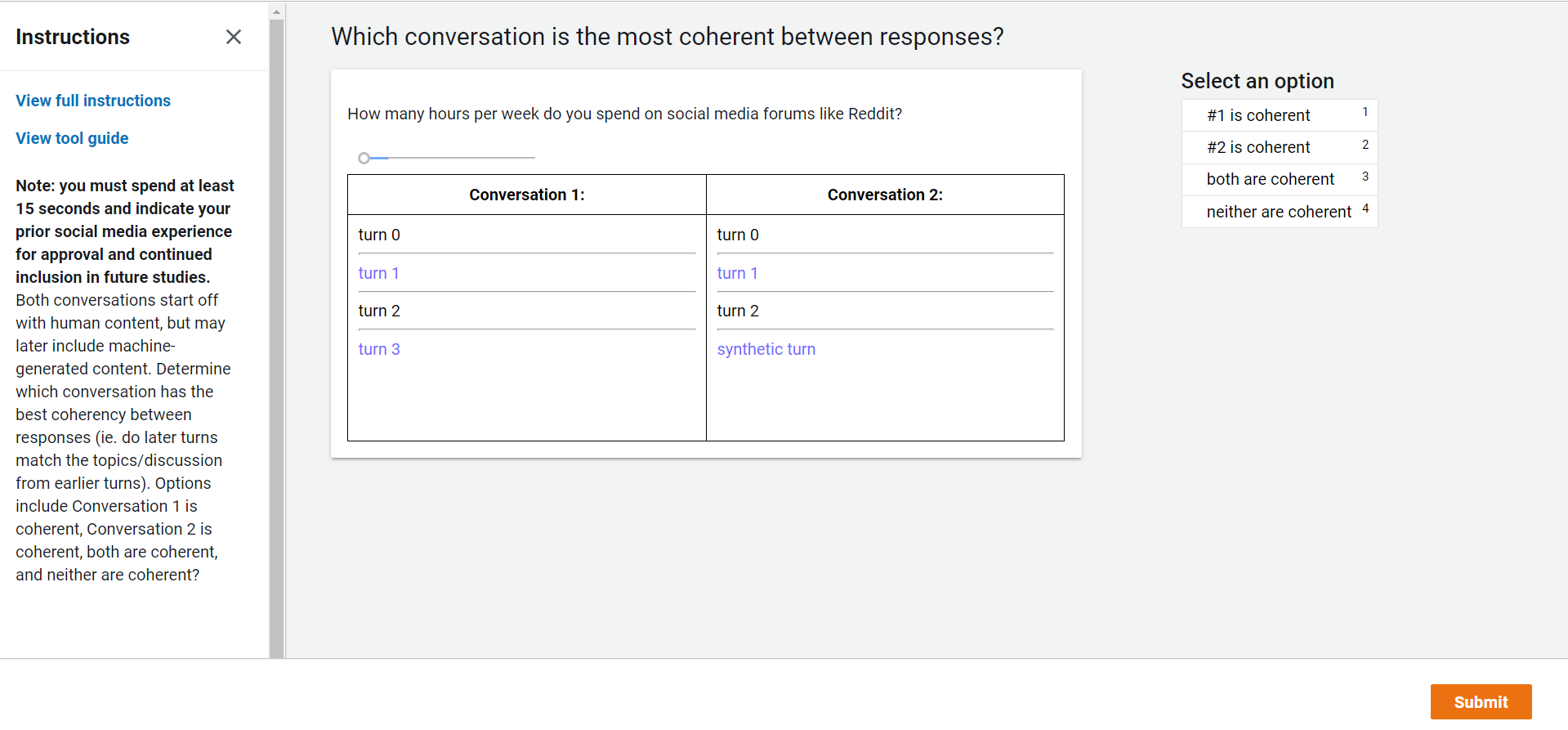}
		\caption{Coherency Task Turk Layout.}
		\label{fig:coh_turk_layout}
	\end{center}
\end{figure*}

Below are sandbox examples of our Mechanical Turk layout for each task as seen in Figures \ref{fig:det_turk_layout}-\ref{fig:coh_turk_layout}. Note that in reality the synthetic example of the pair may come first, or both examples in the pair may be randomly ordered synthetic examples. Also, only the reference task displays the speaker's past reference replies. Hiding the reference in other tasks has the goal of trying to decorrelate the experiments. Example layouts for the realistic, reference, quality, and coherency tests can be found in Figures \ref{fig:det_turk_layout}, \ref{fig:ref_turk_layout}, \ref{fig:qua_turk_layout}, \& \ref{fig:coh_turk_layout} respectively.

\section{Samples}
Listed in Tables \ref{tab:convo_example_2} - \ref{tab:convo_example_8} are three turn generation samples conditioned on varying lengths of real conversations taken from the validation collection. They were all sampled from our Generative Conversation Control model, \decModel(8.3B) using nucleus sampling with $p=0.95$.

All typos present within these samples are intentional and reflective of their sources, be it human generated or sampled. 

\section{Human Evaluation Results}

Listed in Tables~\ref{table:full_references} - \ref{table:full_coherent} are the complete results from our human evaluations, including the percentages that were omitted in Table~\ref{table:human}. The rates that were omitted are two neutral form of ratings: for a given evaluation characteristic (e.g. quality, coherency, etc.) either both sources exhitibited it, or neither did. 

\begin{table*}[h!]
    \centering
    \begin{tabular}{cp{13.5cm}}
    \hline
    Speaker & Reply \\
    \hline \hline
    A    & We're currently investigating reports of players running into issues with connecting to BO4. Can you let us know what region you're in? \\
    & \^ET\\
    \hline
    \emph{B}    & \emph{I’m in the UK and this is what my connection looks like} \\
    \emph{A}    & \emph{Thanks for letting me know, man! We'll be sure to get that escalated. Thanks for your patience! \^WH} \\
    \emph{B}    & \emph{Seems like this is isolated to a select few users, myself included.} \\ 
    \hline
    \end{tabular}
    \caption{A set of three generated responses continuing from the first turn of a real conversation about technical support for a video game between the developer's account and a customer. Upon investigation, it appears that the developer's account signs off their comments with ``\^ET" in order to signal what person in their team wrote the comment. Likewise, it appears the model is emulating this practice as evidenced by signing of ``\^WH".}
    \label{tab:convo_example_2}
\end{table*}

\begin{table*}[h!]
    \centering
    \begin{tabular}{cp{13.5cm}}
    \hline
    Speaker & Reply \\
    \hline \hline
    A    & What’s the best way to run to lose weight? Should I focus on improving my time or increasing my distance? And how should I go about doing either method? Is there another way than just simply pushing myself a little harder each time?\\
    \hline
    \emph{B}    & \emph{What I've found is that it takes a long time to work up a good running program. The best way to lose weight is to control what you eat and to control how many calories you burn. To change the way you eat requires cutting things from your diet. Cutting things on your diet can be incredibly challenging at first but it really comes with time.} \\
    & \emph{It takes several weeks to get used to the new eating habits and to continue eating healthily and exercising.} \\
    & \emph{If you follow a running training program that makes it fun you will want to continue running because of all the good things that come from it. Along with not being overweight (not that that is important to everyone but getting there) I would like to be able to do a half marathon. I don't get to run that much each week but I've found it's a lot easier to run in the cold than in the summer months.} \\
    & \emph{I will say that running helps me with my depression and it's made a big difference in my life in terms of general fitness. If you have nothing else to lose it could really help.} \\
    \emph{A}    & \emph{That’s some solid advice, I’ve definitely known I need to eat less. It’s hard to think of actually budgeting my 
food sometimes though so I’ll keep working at it.} \\
    \emph{B}    & \emph{For me it’s easier to eat under my calories instead of eating an extra 1000. Especially when I’m working on weight loss so my energy levels are a little lower.} \\ 
    \hline
    \end{tabular}
    \caption{A set of three generated responses continuing from the first turn of a real conversation about advice for losing weight.}
    \label{tab:convo_example_3}
\end{table*}

\begin{table*}[h!]
    \centering
    \begin{tabular}{cp{13.5cm}}
    \hline
    Speaker & Reply \\
    \hline \hline
    A    & Which club's transfer window are you most excited about except RM? I'm honestly really curious on how Bayern are gonna do after Lucas Hernandez and Pavard. Is there any rumor about them buying somebody? \\
    B    & Atleast 2 wingers. \\
    & San\'e maybe, but i don't think that City would let him leave anytime soon. P\'ep\'e seems likely, CHO has been rumoured and discussed a lot but due to his very unfortunate injury we won't be approaching him this transfer window. \\
    & Werner is very likely joining us too and its like an ``open" secret at this point. de Ligt has been rumoured a lot but i think that despite the fact that Barcelona already has insane CB Depth, he will end up there.\\
    A    & Brandt? I remember him having an insanely low buy out clause\\
    \hline
    \emph{B}    & \emph{Brandt is totally out of the question. No way he is going to benched for Robben.} \\
    & \emph{Veratti is likely not going to leave aswell. I also think that we won't be buying another Ucl player this winter.} \\
    \emph{A}    & \emph{Say you could choose one to be signed: Werner, San\'e, CHO, Brandt?
} \\
    \emph{B}    & \emph{This is really tough because i love all of them and they all have their own personality/playstyle.} \\ 
    & \emph{Werner is going to be my favorite, maybe because i'm from Germany so this will be a good way for me to support theres local talent.} \\
    & \emph{I love San\'e's self confidence, even though he doesn't really fit us at the moment. With Gnabry, Coman and CHO on the wings it would be hard for him to be more than a rotational player.} \\
    & \emph{CHO seems to be a good player and i would love to see him in Bayern's Shirt but i feel like he will most likely join City this winter.} \\
    & \emph{Brandt is a great player but i fear that he will never displace Robben/Ribery because he doesn't get involved in build up play and he doesn't get into the box which is what we need most on the wings.} \\
    & \emph{I guess I'm going to have to choose between a very good winger that will add a lot to the team and one that will stay put or one who will come in and destroy our team chemistry. I'm torn between the two.} \\
    \hline
    \end{tabular}
    \caption{A set of three generated responses continuing from the first three turns of a real conversation about club soccer players being bought.}
    \label{tab:convo_example_4}
\end{table*}

\begin{table*}[h!]
    \centering
    \begin{tabular}{cp{13.5cm}}
    \hline
    Speaker & Reply \\
    \hline \hline
    A    & If it doesn't appeal to you then don't upgrade. It's subjective and there is nothing to discuss.\\
    B    & Surely all social media platforms are full of subjective posts, hence their popularity?\\
    & I was looking for people in a similar “dilemma” or those that might be able to convince me one way or another, perhaps even expand on a particular aspect of what I posted. \\
    & If you don’t want to contribute an opinion on iPhones then............you know.........don’t ............. \\
    & :D \\
    A    & Probably why I'm not interested in social media.\\
    & I don't have a similar dilemma because I don't find the new iPhone expensive for something I can use everyday for at least 2 years. I have an iPhone X since it came out, I'll probably get the new iPhone this year for bigger RAM and that's it. \\
    \hline
    \emph{B}    & \emph{Fair comment :)} \\
    & \emph{Personally I feel the XS is over priced, for what its got, I’m more than happy with what I got with my SE, I wonder what their update cycle will be as time goes on, surely there will be another notch to fill soon.}\\
    & \emph{I mean, just last year they released a smaller iPhone, and this year they’ve added an even more expensive version,} \\
    & \emph{I mean, I guess if you can afford it, go ahead, but I personally feel you’ve got them hook, line, sinker.} \\
    \emph{A}    & \emph{Yeah, the SE is great.} \\
    & \emph{I don't need the fastest phone, and I didn't get my X for the camera but the wide color gamut, better haptic touch} \\
    & \emph{than any competitor, slightly better display, and RAM and storage upgrades.} \\
    & \emph{The OS is still just as smooth as ever.} \\
    \emph{B}    & \emph{Interesting, I kind of thought that the iPhone X had better brightness (as in almost “daylight” for those of us 
that struggle to see sometimes!)} \\
    & \emph{I appreciate your view!} \\
    \hline
    \end{tabular}
    \caption{A set of three generated responses continuing from the first three turns of a real conversation about upgrading smartphones.}
    \label{tab:convo_example_5}
\end{table*}

\begin{table*}[h!]
    \centering
    \begin{tabular}{cp{13.5cm}}
    \hline
    Speaker & Reply \\
    \hline \hline
    A    & Hi, how hard is it to finish up a T2 set? I am debating between hunter (prinarily for the armor set) vs druid (for gameplay) but I don’t want to roll a hunter and spend all that time to find out T2 full set is not easily achieveable. I’ve never played vanilla before.\\
    B    & It’s the same for Druid. All T2 pieces drop off the same bosses. Hunters are pretty hard though and you’d be expected to kite in the razorgore fight and tranq two other boss fights. But other than that, just do your bow rotation and pump out the best damage you can.\\
    A    & I guess my question was more “is it easier for hunters to get into the raid” since I wouldn’t really care at the pace I go for druid gear but I’d want the T2 set for hunter as soon as possible, but I think that answered my question regardless, thanks!\\
    \hline
    \emph{B}    & \emph{Hunters are obviously the fastest at getting to T2! Some may even be able to get their full set by doing pit of 
Saron and razorgore. That’s assuming you have people who want to beat as many bosses as possible and don’t just rush everything.} \\
    \emph{A}    & \emph{Oof I’m a noob in terms of raid pacing. Yeah I probably don’t have a group that will be able to make it there as soon as possible, but I’ll consider it. Thanks!} \\
    \emph{B}    & \emph{No problem! Hope to see you killing zul if I see you raiding ! (Totally not stalking your guild lmao)} \\
    \hline
    \end{tabular}
    \caption{A set of three generated responses continuing from the first three turns of a real conversation about the video game World of Warcraft.}
    \label{tab:convo_example_6}
\end{table*}

\begin{table*}[h!]
    \centering
    \begin{tabular}{cp{13.5cm}}
    \hline
    Speaker & Reply \\
    \hline \hline
    A    & Putting aside the fact that this ``study" is surely unreliable self-serving garbage, our respective understandings of what can be considered a ``demonstrable fact" are wildly different.\\
    B    & It's only one exeperiment. When you add up all the litany of s ientific evidence,  the conclusion is clear and unequivocal.\\
    A    & ``the conclusion is clear and unequivocal." If you say so.\\
    B    & Provide your evidence that shows otherwise.\\
    A    & You want me to prove a negative? Sure, as soon as you provide evidence that God doesn't exist. (and according to your line of thinking, if you cant provide evidence that he doesn't exist, then he unequivocally does exist).\\
    B    &  You claim it doesn't exist. Provide ypur evidence.\\
    A    & You need to be smarter. Read my last comment and try again. \\
    \hline
    \emph{B}    & \emph{Lol, you provided no evidence to counter anything I said. You're literally just saying your point is right because you said so.} \\
    \emph{A}    & \emph{Dude I gave you an entire anti-theist argument.} \\
    \emph{B}    & \emph{That didn't prove your point at all.} \\
    \hline
    \end{tabular}
    \caption{A set of three generated responses continuing from the first seven turns of a real conversation with two people arguing about the findings of a study.}
    \label{tab:convo_example_7}
\end{table*}

\begin{table*}[h!]
    \centering
    \begin{tabular}{cp{13.5cm}}
    \hline
    Speaker & Reply \\
    \hline \hline
    A    & Bulgaria has a 10\% flat income tax. What does this show?\\
    B    & It shows the income tax you pay in addition to the income tax your employer takes from your salary as well. \\
    A    &  But there is no such thing here, it is 10\% and that's all. That data probably includes all the payments (income tax+health insurance+social payments) taken from your salary, which is another thing.\\
    B    & Social security, pension and health insurance is another 15-20\%. \\
    & That is all considered income tax.\\
    A    & No, it is not. It might be where you live. Here it is not even considered tax, let alone income tax. A tax is something you pay to the state and get nothing in return directly, it is used to keep the state running. You get services for your social payments, so they are not ``taxes".\\
    & Source: I have a law degree\\
    B    & Congrats on the law degree. \\
    & Those are still considered taxes worldwide.\\
    A    & Even if they are that doesn't make the ``income taxes" though. Health insurance for example is owed even if you are unemployed, how is that income tax?\\
    \hline
    \emph{B}    & \emph{Well most employers still charge their employees the 15\% premium for health insurance or even higher.} \\
    & \emph{Your employer deducts 15\% of the salary that isn't declared as income by the employee.} \\
    \emph{A}    & \emph{Again, how is that income tax, but just a deduction.} \\
    \emph{B}    & \emph{Why would it not be considered income tax?} \\
    \hline
    \end{tabular}
    \caption{A set of three generated responses continuing from the first seven turns of a real conversation with two people arguing about what is considered income tax.}
    \label{tab:convo_example_8}
\end{table*}

\begin{table*}[!htb]
\centering\scalebox{0.89}{
\begin{tabular}{lccccr}
\hline 
& Source A better & Both Sources & Neither Source & Source B better & \\ 
Source A & matches Refs. & match Refs. & match Refs. & matches Refs. & Source B \\ 
\hline\hline
\norefModel (355M) & 37.0\% & 19.2\% & 3.0\% & 40.8\% & Human \\
\decModel (355M) & 38.1\% & 19.2\% & 2.9\% & 39.8\% & Human \\
\decModel (774M) & 39.6\% & 21.1\% & 0.8\% & 38.6\% & Human \\
\decModel (1.2B) & 40.1\% & 18.5\% & 1.3\% & 40.0\% & Human \\
\decModel (8.3B) & 41.9\% & 20.0\% & 0.1\% & 37.9\% & Human \\
\hline
\hline
\decModel (355M) & 40.5\% & 19.4\% & 1.2\% & 38.9\% & \norefModel (355M) \\
\decModel (774M) & 38.7\% & 20.0\% & 1.6\% & 39.7\% & \decModel (355M) \\
\decModel (1.2B) & 40.5\% & 20.3\% & 0.9\% & 38.4\% & \decModel (774M) \\
\decModel (8.3B) & 39.3\% & 16.8\% & 0.7\% & 43.2\% & \decModel (1.2B) \\
\hline
\end{tabular}}
\caption{\label{table:full_references} Full results for human evaluations concerning adherence to style in a speaker's reference history (denoted as `Refs.' in the table).}
\vspace{-4mm}
\end{table*}

\begin{table*}[!htb]
\centering\scalebox{0.89}{
\begin{tabular}{lccccr}
\hline 
& Source A is & Both Sources & Neither Source & Source B is & \\ 
Source A & better Quality & are Quality & is Quality & better Quality & Source B \\ 
\hline\hline
\norefModel (355M) & 29.0\% & 32.5\% & 2.9\% & 35.5\% & Human \\
\decModel (355M) & 31.1\% & 32.6\% & 2.8\% & 33.5\% & Human \\
\decModel (774M) & 32.8\% & 30.6\% & 3.2\% & 33.5\% & Human \\
\decModel (1.2B) & 34.4\% & 25.1\% & 2.7\% & 37.8\% & Human \\
\decModel (8.3B) & 41.8\% & 15.3\% & 0.6\% & 42.3\% & Human \\
\hline
\hline
\decModel (355M) & 36.5\% & 24.0\% & 3.4\% & 36.1\% & \norefModel (355M) \\
\decModel (774M) & 33.6\% & 32.7\% & 4.4\% & 29.3\% & \decModel (355M) \\
\decModel (1.2B) & 32.7\% & 32.6\% & 2.4\% & 32.3\% & \decModel (774M) \\
\decModel (8.3B) & 38.2\% & 22.6\% & 0.8\% & 38.4\% & \decModel (1.2B) \\
\hline
\end{tabular}}
\caption{\label{table:full_quality} Full results for human evaluations concerning quality of speech in terms of attributes such as good grammar.}
\vspace{-4mm}
\end{table*}

\begin{table*}[!htb]
\centering\scalebox{0.89}{
\begin{tabular}{lccccr}
\hline 
& Source A is & Both Sources & Neither Source & Source B is & \\ 
Source A & more Realistic & are Realistic & is Realistic & more Realistic & Source B \\ 
\hline\hline
\norefModel (355M) & 31.0\% & 31.5\% & 2.9\% & 34.6\% & Human \\
\decModel (355M) & 32.0\% & 30.9\% & 3.2\% & 33.9\% & Human \\
\decModel (774M) & 31.0\% & 31.2\% & 2.8\% & 35.0\% & Human \\
\decModel (1.2B) & 32.4\% & 27.1\% & 4.0\% & 36.5\% & Human \\
\decModel (8.3B) & 37.4\% & 20.4\% & 1.8\% & 40.4\% & Human \\
\hline
\hline
\decModel (355M) & 30.6\% & 30.1\% & 5.3\% & 33.9\% & \norefModel (355M) \\
\decModel (774M) & 32.7\% & 31.2\% & 3.2\% & 32.9\% & \decModel (355M) \\
\decModel (1.2B) & 30.9\% & 33.8\% & 4.0\% & 31.3\% & \decModel (774M) \\
\decModel (8.3B) & 41.1\% & 21.1\% & 0.9\% & 36.9\% & \decModel (1.2B) \\
\hline
\end{tabular}}
\caption{\label{table:full_realistic} Full results for human evaluations concerning how human-like the conversations were.}
\vspace{-4mm}
\end{table*}

\begin{table*}[!htb]
\centering\scalebox{0.89}{
\begin{tabular}{lccccr}
\hline 
& Source A is & Both Sources & Neither Source & Source B is & \\ 
Source A & more Coherent & are Coherent & is Coherent & more Coherent & Source B \\ 
\hline\hline
\norefModel (355M) & 31.6\% & 26.9\% & 2.5\% & 38.9\% & Human \\
\decModel (355M) & 32.1\% & 29.1\% & 2.7\% & 36.1\% & Human \\
\decModel (774M) & 34.5\% & 25.8\% & 3.3\% & 36.4\% & Human \\
\decModel (1.2B) & 28.6\% & 31.6\% & 4.3\% & 35.6\% & Human \\
\decModel (8.3B) & 40.2\% & 17.0\% & 1.1\% & 41.8\% & Human \\
\hline
\hline
\decModel (355M) & 32.6\% & 27.6\% & 4.9\% & 34.9\% & \norefModel (355M) \\
\decModel (774M) & 34.2\% & 26.3\% & 3.3\% & 36.2\% & \decModel (355M) \\
\decModel (1.2B) & 37.5\% & 22.1\% & 2.7\% & 37.7\% & \decModel (774M) \\
\decModel (8.3B) & 41.5\% & 18.1\% & 1.0\% & 39.3\% & \decModel (1.2B) \\
\hline
\end{tabular}}
\caption{\label{table:full_coherent} Full results for human evaluations concerning how well generated turns fit the conversation as a whole (i.e. coherency).}
\vspace{-4mm}
\end{table*}

\end{document}